\documentclass[a4paper,twoside]{article}

\usepackage{epsfig}
\usepackage{subcaption}
\usepackage{calc}
\usepackage{amssymb}
\usepackage{amstext}
\usepackage{amsmath}
\usepackage{amsthm}
\usepackage{multicol}
\usepackage{pslatex}
\usepackage{apalike}
\usepackage{algorithm2e}
\usepackage[bottom]{footmisc}
\usepackage{url}
\usepackage{comment}
\usepackage{flushend}
\usepackage{geometry}

\usepackage{SCITEPRESS}     % Please add other packages that you may need BEFORE the SCITEPRESS.sty package.

\begin{document}

\title{Content Significance Distribution of Sub-Text Blocks in Articles \\ 
and Its Application to Article-Organization Assessment}

\author{\authorname{You Zhou\sup{1}\orcidAuthor{0009-0005-0919-5793}, Jie Wang\sup{1}\orcidAuthor{0000-0003-1483-2783} \thanks{Correspondence to jie\_wang@uml.edu.}}
\affiliation{\sup{1}Richard Miner School of Computer \& Information Sciences, University of Massachusetts, Lowell, MA 01854, USA}
\email{you\_zhou1@student.uml.edu, jie\_wang@uml.edu}
}


\abstract{We explore how to capture the significance of a sub-text block in an article and how it may be used for text mining tasks.
A sub-text block is a sub-sequence of sentences in the article.
We formulate the notion of content significance distribution (CSD) of sub-text blocks, referred to as CSD of the first kind and denoted by CSD-1.
In particular, 
we leverage Hugging Face's SentenceTransformer to generate contextual sentence embeddings, and use MoverScore over text embeddings to measure
how similar a sub-text block is to the entire text.
To overcome the exponential blowup on the number of sub-text blocks,
we present an approximation algorithm and show that the approximated CSD-1 is almost identical to the exact CSD-1. Under this approximation,
we show that the average and median CSD-1's for news, scholarly research, argument, and narrative articles share the same pattern. We also show that under a certain linear transformation, the complement of the cumulative distribution function of the beta distribution with certain values of $\alpha$ and $\beta$ resembles a CSD-1 curve.
We then
use CSD-1's to extract linguistic features to train an SVC classifier for assessing how well an article is organized. Through experiments, we show that this method achieves high
accuracy for assessing student essays. Moreover,
we study CSD of sentence locations, referred to as CSD of the second kind and denoted by CSD-2, and show that average CSD-2's for different types of articles possess
distinctive patterns, which either conform common perceptions of article structures or provide rectification with minor deviation.
}

%\onecolumn \maketitle \normalsize \setcounter{footnote}{0} \vfill
\maketitle \normalsize \setcounter{footnote}{0} \vfill

\section{\uppercase{Introduction}}
\label{sec:introduction}

In articles crafted by skilled writers, certain sentence positions hold greater significance compared to other positions, as do certain sub-sequences of sentences. A prime example is news articles, where the sentences positioned towards the beginning tend to carry greater significance than those towards the end, resulting in an inverted-pyramid-like structure for content significance. 
In linguistics, article structures are qualitatively classified based on content-significance distributions. Some are classified into self-explanatory geometric shapes, including inverted pyramid, hourglass, diamond, and Christmas tree. Other classifications include narrative, five-boxes, and organic.
The narrative presents a straightforward, chronological account of events. 
For information about five-box and organic structures, the reader is referred to Saleh's guide to article writing \cite{Naveed2014}.

These qualitative classifications serve as a rule of thumb for writers to organize various types of articles and for readers to grab the significant content. However, content significance in an article has not been quantitatively studied, and its full potentials beyond qualitative classification of article structures is yet to be unlocked.

In a recent study on ranking sentences in an article, we note that an ad hoc location weight is assigned to a sentence to reflect its significance based on intuitive judgments specific to the given type of articles \cite{Zhang-Wang2021,Zhang-Zhou-Wang2021}. Despite the rudimentary simplicity of this approach, feature analysis in this study %\cite{Zhang-Wang2021} 
demonstrates that such weights do play a significant role in sentence ranking. 

Motivated by this result, we explore how qualitative classifications of article structures may be turned into quantitative descriptions for carrying out certain text mining tasks. In particular, 
we explore content significance distribution of sub-text blocks in an article that leads to the 
formulation of CSD of the first kind. For this notion to be useful in practice, we need to deal with the exponential blowup of the number of sub-text blocks, for it is time consuming to compute
the exact CSD-1 for a long article. To overcome this obstacle, 
we devise an approximation method to compute CSD-1 over a moderate number of text blocks  
chosen independently at random. We show that, through experiments, the approximated CSD-1 is almost identified to the exact CSD-1.

We investigate four common types of articles: argument, narrative, news, and scholarly research.
To do so, we form four datasets using
existing datasets, one for each type of articles.
We show that the average and median CSD-1 for each type of articles share the same pattern. 
Moreover, we show that a CSD-1 can be resembled as a linear transformation of the complement of the cumulative distribution function of the beta distribution of parameters $\alpha$ and $\beta$ with $0 < \alpha < 1$ and $0 < \beta < 1$. 

To demonstrate the usefulness of CSD-1, we apply CSD-1 to intrinsically determining if an article is well written.  
In particular, we use CSD-1's to extract features and train an SVC classifier using these features to assess intrinsically how well an article is organized. Experiment results show that this method achieves high accuracy
for student essays. 
article organization.
%\end{multicols}

Next, we investigate CSD of the second kind based on sentence locations in an article. Unlike computing CSD-1 that incurs exponential running time, exact CSD-2 can be computed efficiently. We show that the average and median CSD-2's for each type of articles are close to each other, with distinctive patterns for different types of articles, which conform common perceptions of the structures for news and narrative articles, and rectify with minor deviations an earlier perception of the structures for scholarly research and argument articles in the study of sentence ranking \cite{Zhang-Zhou-Wang2021}.

%Through these results we hope to stimulate further interests and investigations on the CSDs of the first and second kinds and possibly CSDs of other kinds yet to be explored.

The rest of the paper is organized as follows: We describe related work on automatic assessment of
article qualities in Section \ref{sec:relatedwork}. We present in Section \ref{sec:preliminary} MoverScore and the datasets.
%for deriving the average and median CSDs of sentence locations and the average and median CSD of text blocks.
In Section \ref{sec:CSD-1} we define CSD of the first kind, describe an approximation algorithm to compute it, and show that CSD-1's for different types of human-written articles all share the same pattern. We also discuss CSD-1 for articles formed by random sentences. In Section \ref{sec:beta}
we show how to resemble a CSD-1 using the beta distribution with $0 < \alpha < 1$ and
$0 < \beta <1$ by the complement of its cumulative distribution function under a linear transformation.
 In Section \ref{sec:grading} we show how to use CSD-1 to intrinsically assess article organization. In Section \ref{sec:CSD-2} we define and discuss CSD of the second kind, denoted by CSD-2. We conclude the paper in Section \ref{sec:conclusion}
 with remarks and suggestions for further investigations.

\section{\uppercase{Related Work}}
\label{sec:relatedwork}

To the best of our knowledge, we are not aware of any prior work on quantitative investigations of content significance distribution of sub-text blocks in an article. 

Automatic assessment of article qualities, on the other hand, has attracted attention in recent years.
The quality of an article is determined by a number of factors, including grammaticality, readability, stylistic attributes, and the depth of expertise presented. 
%Automatic quality assessment has obvious benefits in terms of time savings and tractability in contexts where the volume of documents is large. 
%Various strategies have been devised to facilitate the assessment of document quality across different domains.
Textual analysis can be carried out at the word level, such as identifying verb formation errors, calculating average word frequency, and determining average word length, and at the sentence level. Word-level features have been used to measure word usages and lexical complexity in assessing essays \cite{attali2006automated}. Incorporating sentence-level features for assessing article qualities
provides a fruitful approach \cite{cummins2016constrained}.

In an attempt to address the complexity of assessing article qualities, Yang et al \cite{yang2018automatic} introduced a modularized hierarchical convolutional neural net, where individual sections of an article are treated as separate modules, with an attentive pooling layer applied to the concatenated representations of these sections, which are fed into a softmax layer for evaluation.

To investigate visual effects of an article such as font choices, the layout of the article, and images included in the article,
a multimodal approach \cite{shen2020multimodal} was presented to capture implicit quality indicators that extend beyond the textual content of an article. The integration of these visual aspects with the textual content enhances the effectiveness of article quality assessment.

The concept of text coherence also plays a pivotal role. For example, a hierarchical coherence model was introduced \cite{liao2021hierarchical}, which seeks to leverage local coherence within sentences as well as broader contextual relationships and diverse rhetorical connections. This approach transcends the conventional assessment of sentence similarity, incorporating a richer understanding of the article's coherence.

\section{\uppercase{Similarity measures and datasets}}
\label{sec:preliminary}

It is critical to use an appropriate metric to measure how similar semantically a sub-text block is to the entire text.
Traditional token-based metrics for measuring text similarity, such as BLEU \cite{papineni2002bleu}, ROUGE \cite{lin2004rouge}, and Jaccard's coefficients \cite{Jaccard1912}, fail to capture similarities between texts in lexical forms that convey the same or similar meanings. To overcome this limitation, we would need a metric on semantic similarity.

\subsection{MoverScore}

We propose to use MoverScore \cite{zhao2019moverscore} to measure semantic similarities between two texts. In our case, one text is a sub-text block and the other is the entire article. MoverScore uses Earth Mover's Distance (EMD) \cite{levina2001earth} to compute the distance between the contextual embeddings of the two texts to be compared with,
where contextual embeddings may be computed by ELMo \cite{peters1802deep}, BERT \cite{devlin2018bert}, or some other transformers. 
We use HFace's SentenceTransformer (\cite{reimers2019sentence}) to generate text embeddings.
%
%encode two texts, where EMD is a distance measure that takes into account the semantic similarity between two texts, rather than just their surface forms.
%More specifically, MoverScore uses EMD to compute the distance between the embeddings of the two texts, where the embeddings are computed by weighted sums of word embeddings generated by deep contextual representations such as SentenceTransformer.
%, such as ELMo and BERT, 
%with weights based on inverse document frequency. 
The resulting EMD distance score measures the semantic similarity of the two texts. In so doing, MoverScore provides many-to-one soft alignments to map the candidate word into several most related reference words,
producing a more accurate assessment of the semantic similarities between two texts
% measures the semantic similarity of two texts 
 that is more aligned with human judgment.
%
%
% that incorporates earth mover distance \cite{rubner2000earth} over embeddings, 
%assigning a similarity score to any text block against 
%a reference. 
%It adds earth mover distance \cite{rubner2000earth} over sentence embeddings to 
%MoverScore 

\subsection{Datasets}
We form four datasets, one for each type of articles, using the following existing datasets (all in English): 
SummBank \cite{Radev2003}, Argument Annotated Essays (AAE) \cite{Christian2014}, Predicting Effective Arguments (PEA) \cite{PEA}, BookSum \cite{kryscinski2021booksum}, and research papers from arXiv.
% as datasets for news, argument, and narrative articles. 

\begin{itemize}
\item NewsA for news articles. NewsA is a dataset of 200 news articles selected independently at random from
SummBank, with on average of 23 sentences in an article. SummBank is a large collection of news articles with sentence rankings annotated by three judges.

\item ArguE for argument articles. ArguE is a dataset of 200 essays selected independently at random from the 
union of AAE and PEA,  with an average of 21 sentences in an article.
AAE consists of persuasive essays written by students for preparing for standardized tests, and PEA consists of argument essays written by students of grades 6 -- 12, annotated by experts for discourse elements in argumentative writing.

\item NarrC for narrative articles. NarrC is a dataset of 200 documents selected independently at random from BookSum.
%the following processing, where
BookSum is a large collection of long-form narrative summaries from novels, plays, and stories with a large number of chapter-level documents which covers source documents from the literature domain, such as novels, plays and stories. We randomly select 200 chapters from the datasets which contains 255 sentences each in a chapter. 
%For the selected 200 chapters, we truncate each chapter to around 30 sentences to form NarrC. Our experiments on articles with a larger number of sentences reveal the same features as those of about 30 sentences. Thus, truncating each chapter to around 30 sentences not only can preserve generality and but also can avoid combinatiral blowups.

\item SchRP for scholary research papers. SchRP is a dataset of 200 papers selected independently at random from arXiv.org in Physics, Mathematics, Computer Science, Quantitative Biology, Quantitative Finance, Statistics, Electrical Engineering and Systems Science, and Economics, with approximately an equal number of articles in each subject with an average of 210 sentences in each paper. 
\end{itemize}

\begin{table*}[t]
\centering
\caption{Comparisons between approximated CSD-1 and exact CSD-1 over NewsA} \label{table:0}
\includegraphics[width=0.8\textwidth]{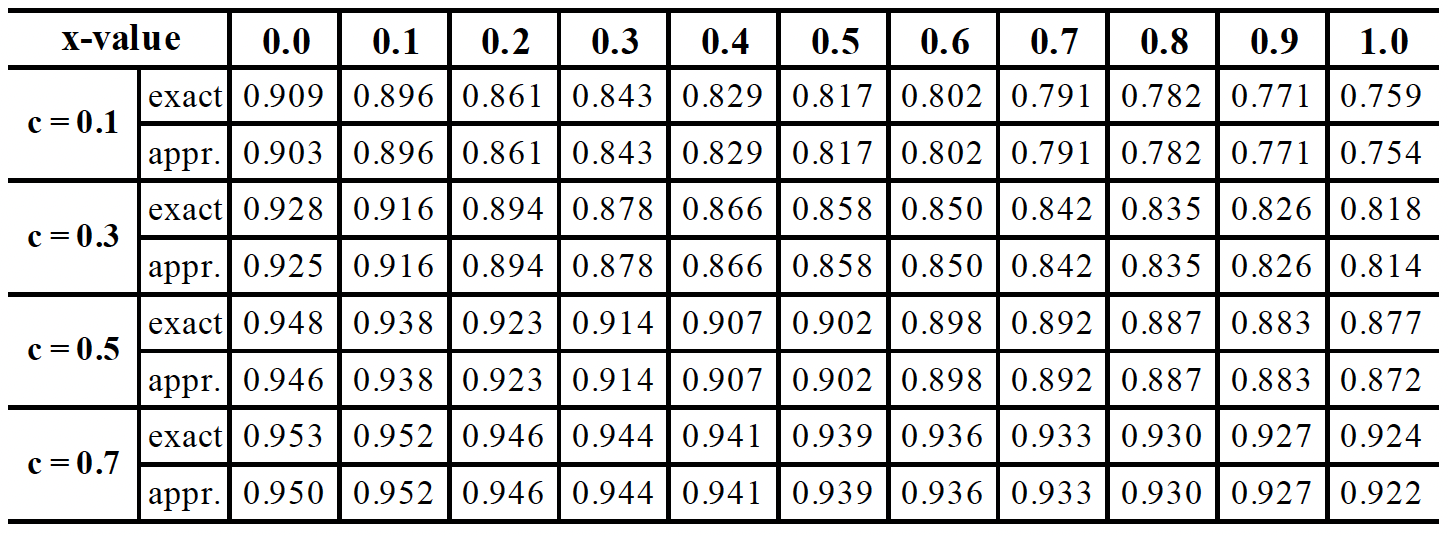}

(a) Average CSD-1 over NewsA

\includegraphics[width=0.8\textwidth]{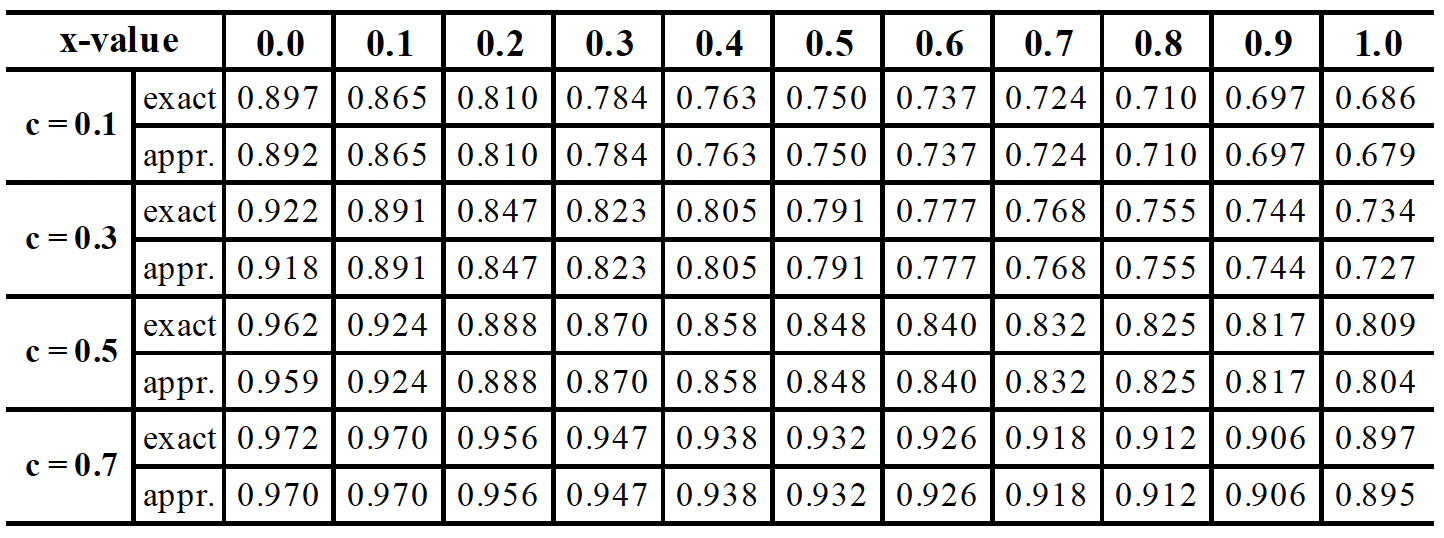}

(b) CSD-1 over a sample news article
\end{table*}

\section{\uppercase{CSD of the First Kind}}
\label{sec:CSD-1}

Let $A = \langle S_1, S_2, \ldots, S_n\rangle$ be an article, represented as a sequence of sentences, where $S_i$ is the $i$th sentence.  Let $k$ be an integer between 1 and $n$.
% and $T_{k,j}$ be the $j$th sub-text block of size $k$ in the lexicographical order or other fixed order. 
There are $N = {n \choose k}$ many sub-text blocks consisting of $k$ sentences. The number of sentences in a text block is also referred to as the size.
For example, let $n = 10$ and $k = \lfloor 0.3n\rfloor =3$. Then
%
%Given an article $D = \langle S_1,S_2,\ldots, S_k\rangle$ of a certain type, a sequence of sentences $S_i$ for $i = 1, 2, \ldots, k$, we compute a MoverScore of each sentence $S_i$ and $D$.
%We consider all sub-sequences of $l = \lfloor 0.3k\rfloor$ sentences of $D$. Such a sub-sequence %is referred to as a text block. For example, if $k = 10$, then $l = 3$, with
there are ${10 \choose 3} =
120$ sub-text blocks of size 3, listed in lexicographical order as follows: $\langle S_1,S_2,S_3\rangle, \langle S_1, S_3, S_4\rangle, \cdots,$ $ \langle S_8, S_9, S_{10}\rangle$.

Let $\text{MSc}(X,Y)$ represent the MoverScore of text $X$ and text $Y$. 
Let $T_1$ and $T_2$ be two text block.
We say that $T_1 > T_2$ if either $\text{MSc}(T_1, A) > \text{MSc}(T_2,A)$ or $\text{MSc}(T_1, A) = \text{MSc}(T_2,A)$ and $T_1$ proceeds $T_2$ in lexicographical order.  
Sort the $N$ sub-text blocks of size $k$ in descending order
according to this ordering and let $T_{k,j}$ be the $j$th sub-text block in the sorted list.
That is, 
$T_{k,1} > T_{k,2} > \cdots > T_{k,N}$.
Then the CSD-1 for $A$ with size $k$ is a discrete function over $x_j = j/N  \in [0,1]$ with $1 \leq j 
\leq N$,
defined by 
$$\text{CSD-1}(A,k,x_j) = \text{MSc}(T_{k,j},A).$$

%We first focus on the patterns of the same type of articles, and then generate 30\% words summary curve to reveal some common patterns between different type of articles.

% Modifying the template --- including but not limited to: adjusting
% margins, typeface sizes, line spacing, paragraph and list definitions,
% and the use of the \verb|\vspace| command to manually adjust the
% vertical spacing between elements of your work --- is not allowed.

% {\bfseries Your document will be returned to you for revision if
%   modifications are discovered.}

%\subsection{Semantic significance distributions}
%We did an experiment on the distribution of key sentences from different types of articles. 
% We derive average CSDs of sentence locations (CSD-SL) and text blocks (CSD-TB) for each type of articles using, respectively, the ArguE, NewsA, NarrC, and SchRP datasets described
% in Section 2.

%We derive the average CSDs of text blocks (CSD-TB) for each type of articles using, respectively, the ArguE, NewsA, NarrC, and SchRP datasets described in Section 2, and articles formed by random sentences, then we show how to use CSD of text blocks to intrinsically evaluate article organization.

% \begin{figure}[!h]
%   \centering
%    {\epsfig{file = img/news1, width = 5.5cm}}
%   \caption{This caption has one line so it is centered.}
%   \label{fig:example1}
%  \end{figure}

%\subsection{CSD of text blocks}

%30\% sentences of the total text is considered that can include most of the information of an article. 
%Given an article $D$ of $k$ sentences, 

We are particularly interested in selecting 30\% of sentences to form a sub-text block because the previous results indicate that selecting 30\% of sentences appropriately from an article would typically capture the major points of the article (see, for example, \cite{Zhang-Wang2021}).

Note that ${n\choose k} \geq (n/k)^k$. For $k = \lfloor 0.3n\rfloor$, we have $(n/k)^k > 3^{0.3n}$, resulting in an exponential blowup. When $n$ is large, computing CSD-1 is intractable. 
For example,
recall that in SchRP of scholarly research papers, the average number of sentences in each article is 210. We have ${n \choose k} = {210 \choose 63} > 3\times 10^{54}$, which is much too big for any computer to handle. 
Approximation is therefore needed. 

\begin{table}[h]
\centering
\caption{Average and median approximated CSD-1 for articles of each type, where the x-values are the normalized sequence of text blocks in ascending order of MoverScores compared with the article itself, and the values in the table are the MoverScores}
\label{table:2}
\includegraphics[width=\linewidth]{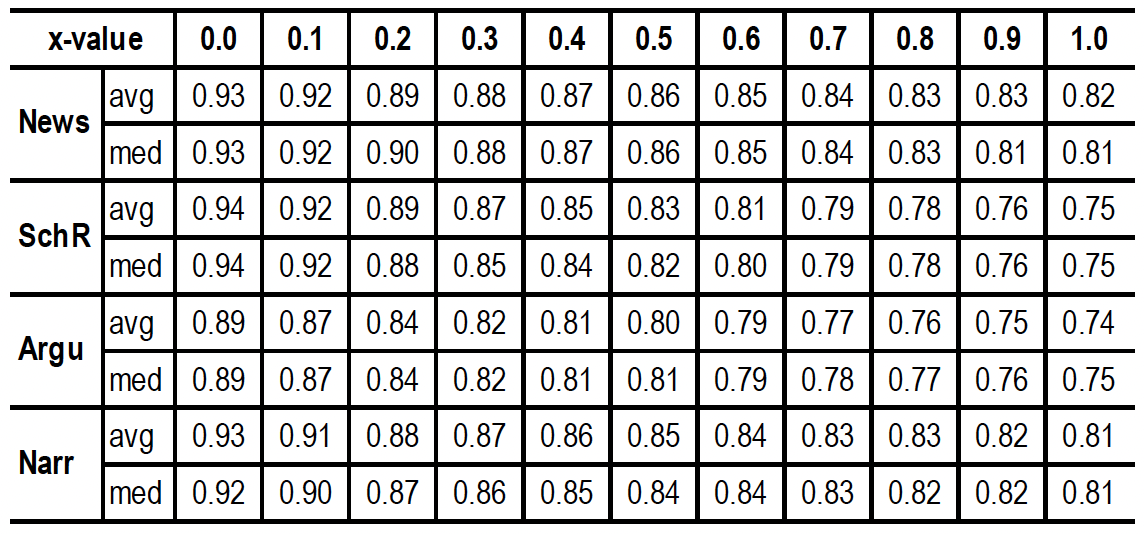}
\end{table}

\subsection{Approximating CSD-1}

Approximating CSD-1 for each article proceeds as follows:

\begin{comment}
%these candidates are generated by choosing the sentences from the original text until it reaches the 30\% sentences limit, we have chosen all the permutations of the sentences as our candidates to get a whole view of the article. Then 
For each text block $T_j$ of $D$, we compute the MoverScore of $T_j$ and $D$. 
%calculate the similarity between candidate and original document, it will return a number between 0 to 1, the larger means the more similar. After 
We sort text blocks in ascending order according to MoverScores and normalize 
the positions of sorted text blocks by $k \choose l$.
Note that when $k = 210$ (recall that in SchRP of scholarly research papers, the average number of sentences in each article is 210), we have ${k \choose l} = {210 \choose 63} > 3\times 10^{54}$, which is much too big for any computer to handle. Approximation is therefore needed. 
We do so for each article as follows:
\end{comment}

\noindent
\begin{figure*}[h] 
	\centering 
	\begin{tabular}{cc}
	\includegraphics[width=0.45\linewidth]{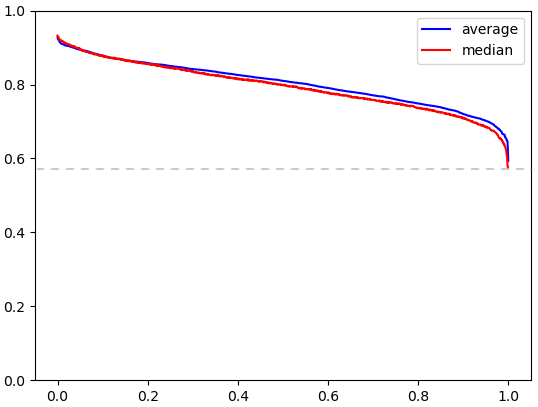} 
	&
	\includegraphics[width=0.45\linewidth]{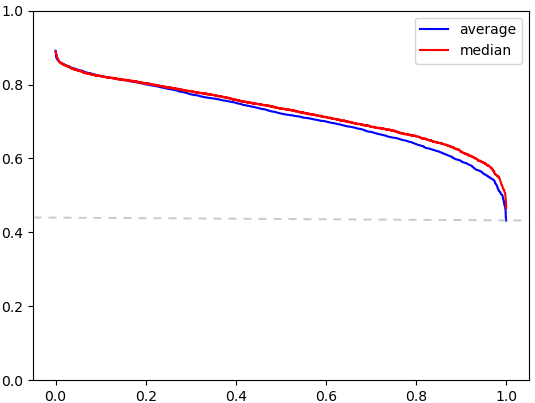} \\
	(a) News & (b) Scholarly research \\
	\includegraphics[width=0.45\linewidth]{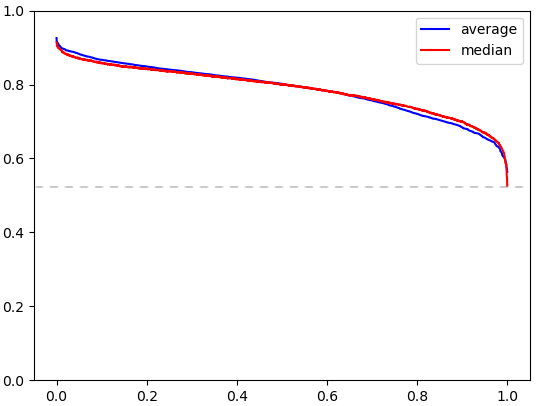} 
	&
	\includegraphics[width=0.45\linewidth]{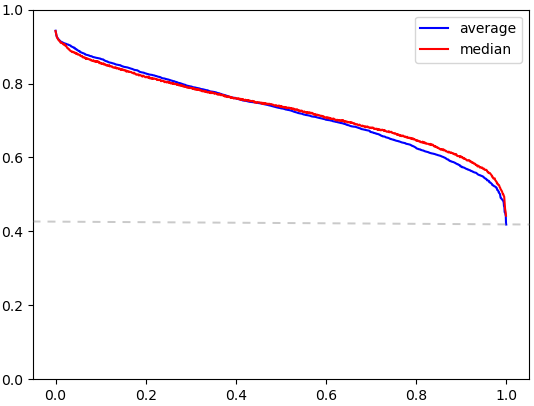} \\
	(c) Argument & (d) Narrative
	\end{tabular} 
	\caption{Average and median approximated CSD-1 for articles of each type, where the x-axis is the normalized sequence of text blocks in ascending order of MoverScores compared with the article itself, and the y-axis is the MoverScores}
	\label{figure:2}
\end{figure*}

\begin{enumerate}
\item 
Generate independently at random 5,000 text blocks. 
%Note that, according to our experiments, choosing 10,000 or more samples produces very similar curves and choosing 5,000 cuts down time complexity significantly. 

\item 
Cluster sentences in $A$ using Affinity Propagation \cite{frey2007clustering} based on 
sentence embeddings generated by SentenceTransformer.
Let $C_1, C_2, \ldots, C_m$ be resulted clusters, where $m$ is determined by the clustering algorithm. Let
$n_i$ be the number of sentences in cluster $C_i$ ($i = 1, 2, \ldots, m$). 
Select $\lfloor 0.3n\rfloor \times n_i/n$ sentences from cluster $C_i$ to form a text block and randomly select 5,000 such text blocks. 

The objective of this step is to avoid pure random sampling and try to cover the exact curve as much as possible with limited samples. We achieve this by
selecting text blocks at random from different topics just in case the text blocks selected in Step 1 have missed certain topics.

\item 
Combine the 5,000 text blocks generated in Step 1 and the 5,000 text blocks generated in Step 2.
Use these 10,000 text blocks to compute CSD-1 as an approximation to the exact CSD-1 for $A$.
\end{enumerate}

Extensive experiments indicate that CSD-1 under this approximation is almost identical to the exact CSD-1 for various sizes. For example, Table \ref{table:0} (a) shows the average approximated CSD-1 and the exact CSD-1 for articles in NewsA, while Table \ref{table:0} (b) shows the approximated CSD-1 and the exact CSD-1 for a random sample from NewsA, where $c$ is a fractional to determine the number $k$ of sentences in a text block with $k = \lfloor cn\rfloor$.
Thus, it is sufficient to use approximated CSD-1 to replace exact CSD-1.

Table \ref{table:2} depicts the average and median approximated CSD-1's with $c = 0.3$ over each dataset,
where SchR, Argu, and Narr represent, respectively, Scholarly Research,  Argument, and  Narrative,
while avg and med represent, respectively, average and median. Figure \ref{figure:2} depicts the corresponding curves.

\begin{comment}
\begin{table}[h]
\caption{Average and median curve data in Figure 2}
\label{} 
\centering
\begin{tabular}{|c|c|c|c|c|c|c|c|c|c|c|c|}
  \hline
   & 0 & 10 & 20 & 30 & 40 & 50 & 60 & 70 & 80 & 90 & 100 \\
  \hline
  news(average) & 0.927 & 0.915 & 0.894 & 0.878 & 0.865 & 0.858 & 0.850 & 0.841 & 0.834 & 0.825 & 0.817\\
  \hline
  news(median) & 0.931 & 0.921 & 0.897 & 0.877 & 0.865 & 0.857 & 0.845 & 0.835 & 0.826 & 0.813 & 0.809\\
  \hline
  narrative(average) & 0.942 & 0.922 & 0.889 & 0.867 & 0.845 & 0.827 & 0.811 & 0.792 & 0.777 & 0.759 & 0.747  \\
  \hline
  narrative(median) & 0.942 & 0.922 & 0.880 & 0.854 & 0.836 & 0.817 & 0.803 & 0.788 & 0.775 & 0.760 & 0.748  \\
  \hline
  scholar(average) & 0.891 & 0.866 & 0.840 & 0.823 & 0.813 & 0.801 & 0.788 & 0.773 & 0.762 & 0.750 & 0.737  \\
  \hline
  scholar(median) & 0.890 & 0.871 & 0.840 & 0.823 & 0.813 & 0.804 & 0.792 & 0.782 & 0.771 & 0.759 & 0.746  \\
  \hline
  argument(average) & 0.926 & 0.906 & 0.882 & 0.867 & 0.857 & 0.849 & 0.840 & 0.833 & 0.826 & 0.818 & 0.810  \\
  \hline
  argument(median) & 0.916 & 0.900 & 0.872 & 0.859 & 0.849 & 0.842 & 0.835 & 0.829 & 0.822 & 0.815 & 0.807  \\
  \hline
\end{tabular}
\end{table}

Figure \ref{figure:2} depicts the average and median CSDs of text blocks for randomly selected 200 articles from each dataset.
\end{comment}

\subsection{Segments of CSD-1}

\begin{figure}[h]
\centering
\includegraphics[width=0.93\linewidth]{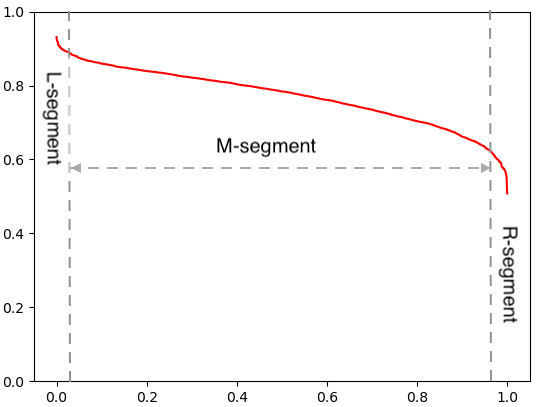}
\caption{CSD-1 segments}
\label{fig:segment}
\end{figure}

It is evident that the average approximated CSD-1 and the median approximated CSD-1 for each type of articles are very close to each other, and they all share the same pattern.
This pattern can be divided into three segments, referred to as, from left to right, \textsl{the left segment} (L-segment), \textsl{the middle segment} (M-segment), and \textsl{the right segment} (R-segment). The L-segment contains a very small number of text blocks that are substantially more significant than the rest, with the largest value close to 0.9. The union of these text blocks is the most significant content of the article. The R-segment contains a very small number of text blocks with the least significance, with value below 0.6 and above 0.4. These text blocks often contain connection sentences. The M-segment is the majority of text blocks that are gradually decreasing in terms of significance. Figure \ref{fig:segment} depicts the three segments of CSD-1. 

\subsection{CSD-1 for randomly generated articles}

To establish a baseline for CSD-1 on articles written by under-educated writers, we generate articles by selecting at random unrelated sentences from 20 existing articles, one sentence per article, and placing them in a random order. We call such articles ``Random sentences". Existing articles are selected from SummBank in one setting, and Wikipedia \cite{wikidump} in another setting.

% \begin{comment}
% \begin{figure}[h] 
% 	\centering 
% 	\includegraphics[width=\linewidth]{img/random_human_compare.png} 
% 	\caption{Comparison between human article and random article curve} 
% 	\label{figure:position}
% \end{figure}
% \end{comment}

% 相似曲线
% In order to support our argument, we conducted two additional experiments. Firstly, we selected 20 identical sentences and 20 sentences with high embedding similarity to form an article. In the case of the 20 identical sentences, we observed that all curve percentages appeared as flat straight lines. Furthermore, to generate similar sentences, we replaced only one or two words with synonyms, where their embedding similarity was greater than 0.9. And the 30\% curve for the 20 similar sentences comparing with a human write article is shown below. When compared to the curve generated by human-written articles, it is evident that the distribution of curves for similar sentences on the Y-axis is significantly shorter. The minimum similarity is approximately 0.8, indicating that regardless of the selected combination of sentences, they are all closely related to the original article. But a normal article written by individual can not reach this high similarity, although all sentences in an article should revolve around a single theme, their text similarity would not be so close, especially considering that some sentences are transitional and lack actual meaning.

\begin{figure}[h] 
	\centering 
	%\begin{tabular}{cc}
	\includegraphics[width=0.93\linewidth]{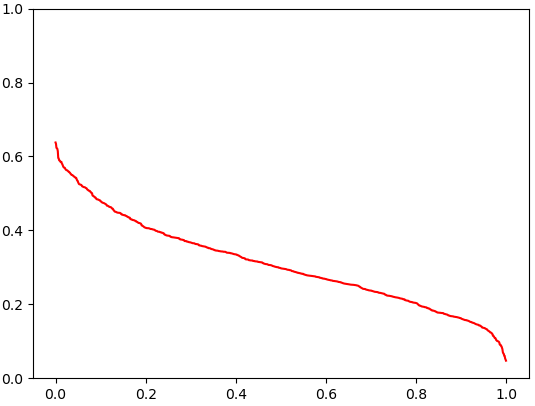} \\
	(a) Random sentences \\
    \includegraphics[width=0.93\linewidth]{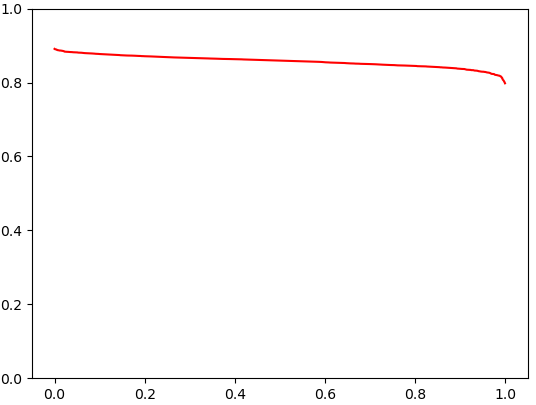} \\
	(b) Similar sentences
	%\end{tabular}
	\caption{Average CSD-1's with $c = 0.3$ for (a) articles of random sentences and (b) articles of similar sentences, with the same x-axis and y-axis as those in Figure \ref{figure:2}} 
	\label{figure:3}
\end{figure}

We also form articles using 20 identical sentences and 20 sentences with high embedding similarity. For the former, each CSD-1 is simply a straight line. For the latter, we generate similar sentences from a given sentence as follows:  Select at random one or two words and replace them with synonyms to form 20 new sentences so that their pairwise embedding similarity is greater than 0.9. 
Figure \ref{figure:3} depicts the average CSD-1 on random-sentence articles and
similar-sentence articles. It can be seen that the average CSD-1 for random-sentence articles exhibits a much larger y-value range from nearly zero to below 0.7. 
%Text blocks with nearly zero MoverScores indicate that they are irrelevant to the article as a whole. 
%The average CSD-TB for similar-sentence articles has a minimum value greater than 0.8, which is substantially greater than human-written articles as shown in Figure \ref{figure:2}.

% \begin{comment}
% \begin{figure}[h] 
% 	\centering 
% 	\includegraphics[width=\linewidth]{img/sim_human_compare.png} 
% 	\caption{Comparison between human article and similar sentences article curve} 
% 	\label{figure:position}
% \end{figure}
% \end{comment}

\section{\uppercase{Transforming Beta Distribution to CSD-1}}
\label{sec:beta}

We show that a CSD-1 curve can be resembled using the beta distribution for certain values of parameters $\alpha$ and $\beta$ under a certain linear transformation. In particular, we present the following observations:
\begin{enumerate}
\item A typical CSD-1 curve resembles the complement of a cumulative distribution function (CDF) 
of a U-shape probabilistic density function (PDF).

\item The beta distribution, denoted by 
$\text{Beta}(\alpha, \beta)$, provides a U-shape PDF with $0< \alpha < 1$ and $0 < \beta < 1$.
\end{enumerate} 

\begin{figure}[h]
\centering
\includegraphics[width=0.93\columnwidth]{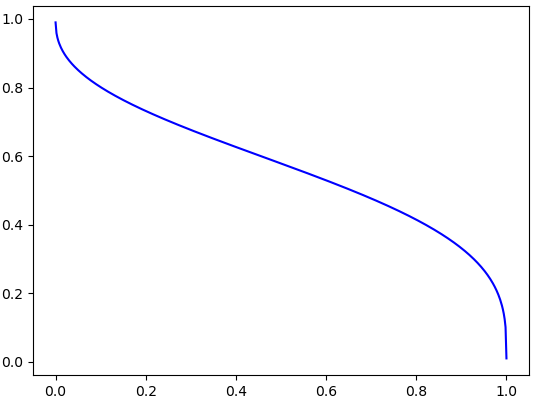}
\caption{The curve of $C_x(0.4,0.3)$, which spans the entire y-axis from 0 to 1}
\label{fig:cdf}
\end{figure}

Denote by $I_x(\alpha, \beta)$ the CDF of $\text{Beta}(\alpha, \beta)$. Let
$$C_x(\alpha, \beta) = 1 - I_x(\alpha, \beta).$$
Figure \ref{fig:cdf} shows the curve of $C_x(0.4, 0.3)$. 
 
We apply a linear transformation to obtain a CSD-1 curve in the desired range. Let
$$LC_x(a,b \mid \alpha,\beta) = a\cdot C_x(\alpha, \beta) + b,$$
where $a \geq 0, ~b\geq 0$, and $a+b \leq 1$. For easier reading, we may also write
$LC_x(a = a_0, b=b_0 \mid \alpha = \alpha_0,\beta=\beta_0)$ as $LC_x(a_0,b_0\mid \alpha_0,\beta_0)$.

\begin{figure}[thb]
\centering
\includegraphics[width=0.93\columnwidth]{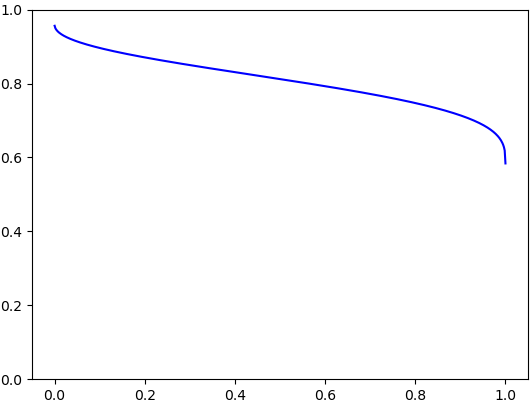} \\
(a) $LC_x(0.38,0.55\mid 0,45,0.3)$ \\
\vspace{4pt}
\includegraphics[width=0.93\columnwidth]{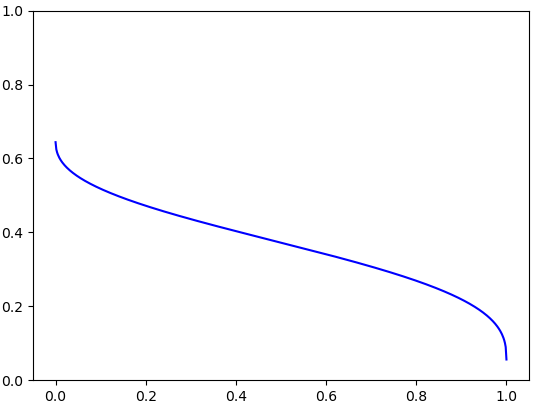} \\
(b) $LC_x(a = 0.6, b=0.05 \mid \alpha=0,4, \beta=0.35)$ \\
\vspace{4pt}
\includegraphics[width=0.93\columnwidth]{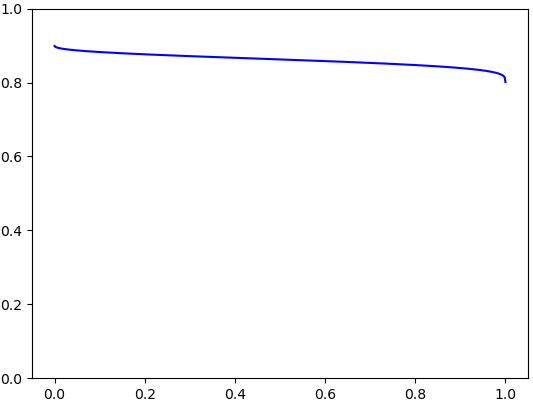} \\
(c) $LC_x(a = 0.1, b=0.8 \mid \alpha=0,4, \beta=0.25)$ \\
\caption{The curves resemble, respectively, (a) the average and median curves for news articles in Figure \ref{figure:2}(a), (b) the average curve for random sentences in Figure \ref{figure:3}(a), and (c) the average curve for similar sentences in Figure \ref{figure:3}(b)}
\label{fig:LC}
\end{figure}

For example, Figure \ref{fig:LC}(a) depicts the curve of $LC_x(a = 0.38, b=0.55\mid \alpha=0,45, \beta=0.3)$, which resembles the curves of Figure \ref{figure:2}(a) for NewsA. 
Figure \ref{fig:LC}(b) depicts the curve of $LC_x(a = 0.6, b=0.05 \mid \alpha=0,4, \beta=0.35)$, which resembles the curves of Figure \ref{figure:3}(a) for random sentences. 
%
%\begin{figure}[h]
%\centering
%\includegraphics[width=0.93\columnwidth]{img/Random.png}
%\caption{The curve of $LC_x(a = 0.6, b=0.05 \mid \alpha=0,4, \beta=0.35)$ that resembles the average curve for random sentences in Figure \ref{figure:3}(a)}
%\label{fig:random}
%\end{figure}
%
Figure \ref{fig:LC}(c) depicts the curve of $LC_x(a = 0.1, b=0.8 \mid \alpha=0,4, \beta=0.25)$, which resembles the curves of Figure \ref{figure:3}(b) for similar sentences. 

%\begin{figure}[h]
%\centering
%\includegraphics[width=0.93\columnwidth]{img/Similar.png}
%\caption{The curve of $LC_x(a = 0.1, b=0.8 \mid \alpha=0,4, \beta=0.25)$ that resembles the average curve for similar sentences in Figure \ref{figure:3}(b)}
%\label{fig:similar}
%\end{figure}

We observe that $\alpha$ controls how the L-segment of the CSD-1 curve looks like and $\beta$ controls how the R-segment of the CSD-1 curve looks like. In particular, 
\begin{enumerate}
\item If $\alpha$ is smaller, then the L-segment is larger both vertically and horizontally.
\item If $\beta$ is smaller, then the R-segment is larger both vertically and horizontally.
\end{enumerate}

We conjecture that by choosing the values of $\alpha$, $\beta$, $a$, and $b$ appropriately, we can resemble CSD-1 for any type of article using $LC_x(a,b\mid \alpha, \beta)$.

\section{\uppercase{Assessing article organization}}
\label{sec:grading}

% Each author must be defined separately for accurate metadata
% identification. Multiple authors may share one affiliation. Authors'
% names should not be abbreviated; use full first names wherever
% possible. Include authors' e-mail addresses whenever possible.

% Grouping authors' names or e-mail addresses, or providing an ``e-mail
% alias,'' as shown below, is not acceptable:
% \begin{verbatim}
%   \author{Brooke Aster, David Mehldau}
%   \email{dave,judy,steve@university.edu}
%   \email{firstname.lastname@phillips.org}
% \end{verbatim}

% The \verb|authornote| and \verb|authornotemark| commands allow a note
% to apply to multiple authors --- for example, if the first two authors
% of an article contributed equally to the work.

% If your author list is lengthy, you must define a shortened version of
% the list of authors to be used in the page headers, to prevent
% overlapping text. The following command should be placed just after
% the last \verb|\author{}| definition:
% \begin{verbatim}
%   \renewcommand{\shortauthors}{McCartney, et al.}
% \end{verbatim}
% Omitting this command will force the use of a concatenated list of all
% of the authors' names, which may result in overlapping text in the
% page headers.

% The article template's documentation, available at
% \url{https://www.acm.org/publications/proceedings-template}, has a
% complete explanation of these commands and tips for their effective
% use.

% Note that authors' addresses are mandatory for journal articles.

As an application of CSD-1, we show how to assess intrinsically how well an article is organized using features extracted from multiple CSD-1's with various sizes. For this purpose, we use the ASAP (the Automated Student Assessment Prize ) dataset \cite{Christian2014},
% of $l_i = \lfloor (0.1ik\rfloor$, where
%$i = 1, 2, \ldots, 9$. 
for it contains subsets suitable for carrying out this task. It consists of eight sets of essays, written by students and scored by experienced graders on idea, organization, style, convention, sentence fluency, and word choice. 

In particular, we choose Set 7 and Set 8, for they provide an organization score for each essay, assessing whether an essay is organized in a way that enhances the central ideas and its development, focusing on whether the sentence orders are compelling and move the reader through the text easily, and the connections between ideas and events are clear and logically sequenced. 
%Based on the description, this score mostly reflects the article structure which can be measured by the distribution of key sentences we mentioned above.

%\textbf{Datasets and measures}
Specifically, Set 7 consists of 1,730 essays with an average of 250 words, scored by two graders with scores from 0 to 3. 
Averaging the two scores for each essay yields five labels from 1 to 3 with an increment of 0.5, one essay one score. 
Set 8 consists of 918 essays with an average of 650 words, scored by two or three graders with scores from 1 to 6. Averaging scores for an essay may result in slight discrepancy depending on whether it has two or three scores. For example, an average score could be 3.5 with 2 scores or 3.67 with 3 scores for different papers of about the same quality. We combine such scores into one score, yielding eleven labels from 1 to 6 with an increment of 0.5.

\begin{figure}[!h]
  \centering
  \includegraphics[width=0.93\linewidth]{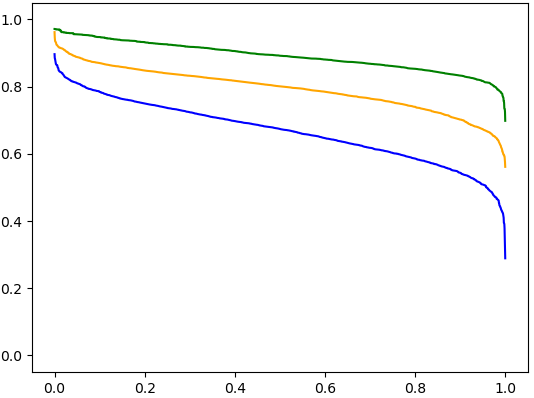}
  % {\epsfig{file = img/20-50-80-reverse.png, width = 5.5cm}}
  \caption{Approximated CSD-1's with sizes determined by $c_i = 0.2+3(i-1)/10$ for $i=1, 2, 3$.}
  % $L_1$ (the lower curve), $L_2$ (the middle), and $L_3$ (the upper) for a single document in Set 7, where $L_i = \lfloor (0.2+3(i-1)/10)n\rfloor$ for $i=1, 2, 3$.}
  \label{fig:example1}\label{figure:4}
\end{figure}

%For privacy, names of people and places, numbers, and dates in the raw data are replaced, respectively, with place holders @person, @location, @num and @date, we randomly generate names, numbers and dates to replace them in the raw data to get a better understanding of the article.
Neither set provides sufficient data to train neural-net classifiers. Instead, we train SVC classifiers by extracting features from approximated CSD-1 of various sizes as percentages of $n$, the number of sentences in an article. For example, we may divide $n$ by 10 or by 5 to
yield nine or 19 different sizes.
% of $\lfloor 0.1ik\rfloor$ for $i = 1, 2, \ldots, 9$.
%, or divide 10 by 5 to yield 19 different sizes of $\lfloor 0.05ik\rfloor$ for $i = 1, 2, \ldots, 19$.
Figure \ref{figure:4} shows approximated CSD-1's of text blocks of three different sizes of a single document in Set 7,
which are of essentially the same shape with the one of a larger size above the one of a smaller size.

However, even with 10,000 text blocks to obtain an approximated CSD-1, the computation is still too high, which hinders our application of assessing essay organization in real time. To overcome this obstacle, we sample $N$ text blocks uniformly and independently at 
random to further approximate CSD-1's on each size.
Through intensive and extensive experiments,
we find that choosing $N = 1,000$ and dividing $n$ by 10
offer a satisfactory trade-off between accuracy and time complexity for our applications.
Note that selecting a larger value of $N$ and dividing $n$ by a
number smaller than 10, while providing a better accuracy, incurs a tremendous time complexity required to calculate CSD-1's for 9 times from 10\% to 90\% for each article, 
making it unpractical.
This yields, for each article, nine vectors of 1,000 dimensions % and 19 2,000-d vectors 
from the corresponding 
CSD-1. We then select 10 equal positions from a 1,000d vector to produce, for each article, nine 10d vectors.
% for setting 1 and 19 10-d vectors for setting 2. Likewise, we select 10 equal positions from a 2,000-d vector to form two 10-d vector, yielding nine 10-d vectors for setting 1 and 19 10-d vectors for setting 2. 

We train a multi-label SVC classifier for Set 7 based on the given training set using the 10d feature vectors with a standard 80-20 split, and do the same for Set 8. 
%denoted by $C_{n,i}$, where $n$ is the number of text blocks sampled and $i$ indicates which setting. Thus, for Set 7, we obtain four SVC classifiers: $C_{1,000, 1}$, $C_{1,000, 2}$, $C_{2,000,1}$, and $C_{2,000,2}$. 
%Their meanings are self explanatory.

For a test article, we call the predicted result ``exact", ``adequate", and ``acceptable" if, respectively, the predicted label by SVC is identical to its label, $\pm 0.5$ of the exact,
and $\pm 1$ of the exact.
It is common for experienced graders to have small dependencies among them, with $\pm 0.5$ been adequate and $\pm 1$ acceptable. Table \ref{table:1} shows the binary F1 scores
for ``exact", ``adequate", and ``acceptable". 

% \begin{comment}
% \begin{table}[h]
%     \caption{Comparison results(\%) on Set 7 and 8 with sampling 1000 and 2000, features extracting from 10\%-90\% and 5\%-95\%}
%     \centering
%     \begin{tabular}{|c|r|r|r|r|}
%         \hline
%        % \multirow{}{} 
%         &
%         \multicolumn{2}{c|}{Sampling 1000} & \multicolumn{2}{c|}{Sampling 2000}\\
%         \cline{2-5} 
%         & 10\%-90\% & 5\%-95\% & 10\%-90\% & 5\%-95\% \\ \hline 
%         Essay set 7 & 71.26 & 72.03 & 73.52 & 75.03 \\ \hline 
%         Essay set 8 & 67.19 & 70.90 & 72.11 & 73.65 \\ \hline 
%     \end{tabular}
%     \label{tab:my_label}
% \end{table}
% \end{comment}

% \begin{table}[h]
% \vspace{-0.2cm}
% \caption{This caption has more than one line so it has to be
% justified.}
% \label{tab:example2} 
% \centering
% \begin{tabular}{|c|c|}
%   \hline
%   Example column 1 & Example column 2 \\
%   \hline
%   Example text 1 & Example text 2 \\
%   \hline
% \end{tabular}
% \end{table}

\begin{table}[h]
\caption{F1 scores of predicted labels for test articles}
\label{table:1}
\centering
\begin{tabular}{l|c|c|c}
\hline
       & Exact (\%) & $\pm 0.5$ (\%) & $\pm 1$ (\%) \\\hline
Set 7  & 71.26 & 85.21 & 99.08 \\\hline
Set 8 & 67.19 & 76.66 & 98.01 \\\hline
\end{tabular}
\end{table}

In developing a practice application, for datasets with a smaller number of labels, such as Set 7 with 5 labels, we may hesitate to consider $\pm 1$ of the exact acceptable, but it is reasonable to consider $\pm 1$ of the exact acceptable for
 datasets with a larger number of labels, such as Set 8 with 11 labels. Thus, we would want to adopt the 6-point system to achieve over 98\% accuracy of acceptable intrinsic evaluation of article organization.

%It can be seen that sampling 2000 achieves a better accuracy compared with sampling 1000 because doing sampling means lost a lot of information to gain speed, one thousand out of hundred thousand means that it can not promise to cover a whole view of the curve. Although sampling 2000 is still not enough, it can cover the curve more thoroughly. Meanwhile, extracting features every 5\% get a higher accuracy than every 10\%. It is obvious that features extracted every 5\% have is more than every 10\%, which achieved a higher accuracy. And on the comparison, it improves more especially in the Essay set 8. Because the average length essay set 8(average 35 sentences) is longer than essay set 7(average 20 sentences), for those essays which length is less than 15 sentences, features every 5\% would not introduce many effective features as some features at some percentage might be the same, these features are redundant during classification. While for essays more than 20 sentences, every 5\% features contribute 3 times as many features as every 10\%, which leads to a higher accuracy.

\section{\uppercase{CSD of the second kind}}
\label{sec:CSD-2}

We define CSD-2 to reflect the significance of sentence locations. The idea is to identify where the top 30\% of sentences with the highest MoverScores are located. The reason of choosing 30\% is the same as that in Section \ref{sec:CSD-1}.
%and compute average and median CSD-2's for each type of articles over, respectively, the ArguE, NewsA, NarrC, and SchRP datasets.

Given an article $A = \langle S_1,S_2,\ldots, S_n\rangle$ of a certain type with $S_i$ being the $i$th sentence for $i = 1, 2, \ldots, n$, 
%we compute $\text{MSc}(S_i, A)$, a MoverScore of each sentence $S_i$ and $A$. 
% trying to having a whole view on which part of an article have a higher similarity of different types of articles. We calculate the similarity between all sentences and the original text to find 
select the top $t = \lfloor 0.3n\rfloor$ sentences that are most similar to $A$ as key sentences determined by MoverScores. Let these sentences be $S_{i_1}, S_{i_2},\ldots, S_{i_t}$ with
$i_1 < i_2 < \cdots < i_t$.
We normalize the location index of each of these sentences in $A$ by $n$ and define CSD-2 as the following discrete function:
\[
\text{CSD-2}(A,i/n) = 
\left\{
\begin{array}{ll}
\text{MSc}(S_{i_j},A), & \mbox{if $i = i_j$}, \\
0, & \mbox{otherwise.}
\end{array}
\right.
\]

\begin{figure*}[t] 
	\centering 
	\begin{tabular}{cc}
	\includegraphics[width=0.45\linewidth]{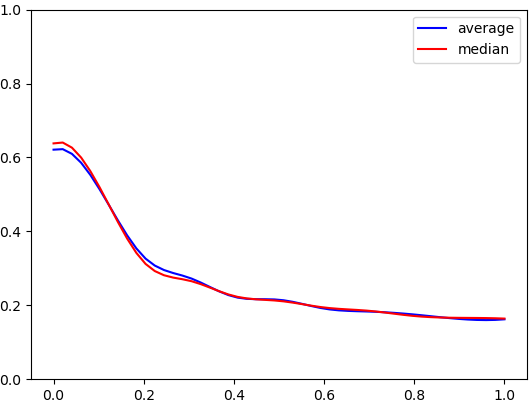} 
	&
	\includegraphics[width=0.45\linewidth]{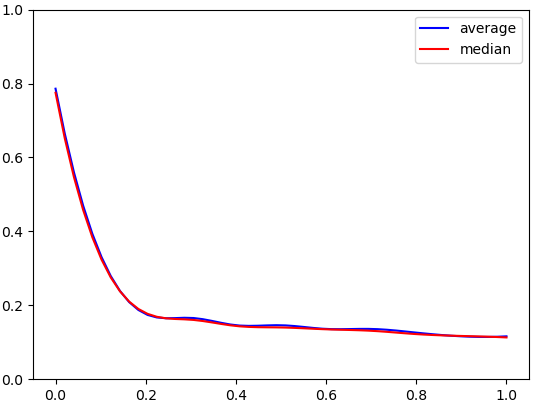} \\
	(a) News & (b) Scholarly research \\
	\includegraphics[width=0.45\linewidth]{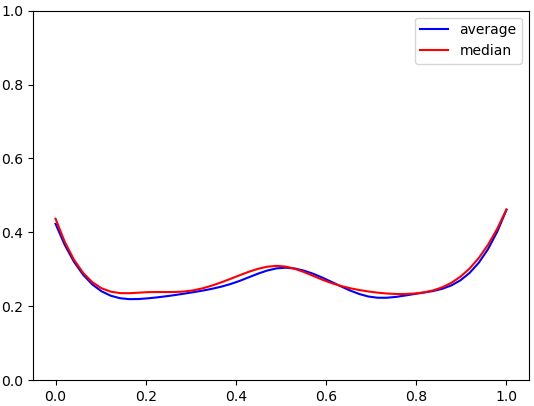} 
	&
	\includegraphics[width=0.45\linewidth]{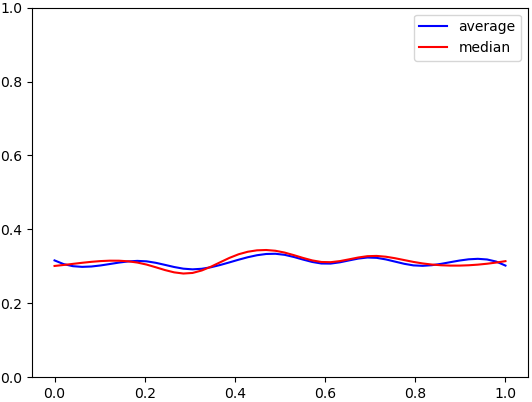} \\
	(c) Argument & (d) Narrative
	\end{tabular} 
	\caption{Average and median CSD-2 for articles of each type, where the x-axis is the normalized sentence indexes and the y-axis is the MoverScores of sentences and the article they are in}
	\label{figure:1}
\end{figure*}

We compute the average and median CSD-2's for each type of articles over, respectively, the ArguE, NewsA, NarrC, and SchRP datasets.
%Averaging the 200 articles in each dataset to obtain 
%the average CSD-2 for these articles.
%We also look at the median CSD-2.
Table \ref{table:1} depicts the average and median CSD-2 for each dataset,
%where SchR, Argu, and Narr represent, respectively, Scholarly Research,  Argument, and  Narrative,
% avg and med represent, respectively, average and median.
and Figure \ref{figure:1} depicts the corresponding CSD-2 curves.
% for the four datasets.
%The reason to choose top 30\% of sentences is that they typically capture all the major points of an article.
%

\begin{table}[h]
\caption{Average and median CSD-2 for each dataset}
\label{table:1} 
\centering
\includegraphics[width=\linewidth]{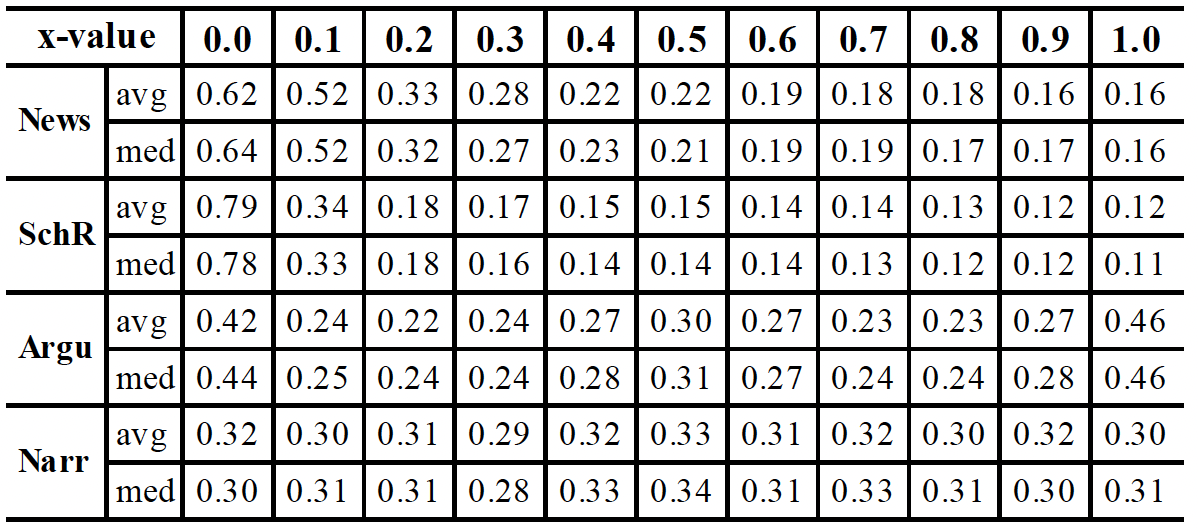}
\end{table}

\noindent

The following observations are evident. 
\begin{enumerate}
\item For each dataset, the median CSD-2 is very closed to the average CSD-2, indicating that the average CSD-2 curve provides a good representative for each dataset. 

\item 
For news articles, the average CSD-2 is monotonically decreasing. 
This is in line with an inverted pyramid in the common perception for the structure of news articles.
The small bumps over the x-interval $(0.2, 0.6)$ indicate that the structure of a news article may not be a straight inverted pyramid. 

\item 
For scholarly research articles, the average CSD-2 is also monotonically decreasing. Different from the average CSD-2 for news articles, we note that the values over the x-interval $(0,0.2)$ drops rapidly from about 0.8 to below 0.2. It decreases slowly over the x-interval $(0.2, 1]$. This indicates that in a scholarly research paper, the first 20\% of sentences would be the most significant, and the rest of the article would be justifications of these statements. This differs from an earlier perception that presumes an hourglass structure for scholarly research articles \cite{Zhang-Wang2021}.
%This differs from a previous imagination that the structure of scholarly research articles is like a hourglass \citep{Zhang-Zhou-Wang2021}.
%
%
%It is clear that in the news, most of the important sentences are at the beginning. There is a four “W” rule in writing a piece of news: you need to establish the “who”, “what”, “when”, and “where” of your piece. The “why” and “how” can wait until the following paragraphs. A journalism piece looks like a pyramid. The most important information goes at the top. The rest is spread throughout the remaining column space. It is logical that most news always states the event at first and then describes more details after, so the key sentences are those statement sentences.
%
\item 
For argument articles, the average CSD-2 resembles a shallow ``W'', with two peaks at both ends and one peak in the middle, where the peaks at both ends are higher than the one in the middle.
This indicates that the structure of an argument article would, on average, start and end with the most significant arguments that match each other, with other secondary arguments somewhere in the middle. This differs from an earlier perception that presumes a pyramid structure for argument articles \cite{Zhang-Wang2021}.
\item
For narrative articles, the average CSD-2 resembles a shallow wave line centered around a value slightly greater than 0.3. This indicates that the structure of a narrative article would, on average, includes sentences of approximately the same significance throughout the article. This is  in line with the common perception.
\end{enumerate}

\section{\uppercase{Conclusions and Final Remarks}}
\label{sec:conclusion}

We present for the first time a quantitative method to capture content significance distributions of sub-texts within an article and show that it is a promising new approach to unlocking potentials for certain text mining tasks using linguistic knowledge which so far only has qualitative descriptions. 
In addition to CSD-1 and CSD-2, we believe that CSDs of other kinds may also be possible that are awaiting to be explored. 

We have demonstrated how to use CSD-1 to assess article organization with high accuracy, and we believe that other applications may also be possible.
For instance, we may use CSD-2 to identify the type of a given article to help obtain more accurate ranking of sentences. Incorporating sentence ranking to a large language model such as GPT-3.5-turbo \cite{brown2020language}, LLaMA \cite{touvron2023llama}, and PaLM \cite{chowdhery2022palm} is expected to help generate a better summary for a given article.

Our approach of computing CSDs relies on metrics of comparing semantic similarities of a sub-text block (note that a sentence is a special case of sub-text block) to the article it is in. While MoverScore is arguably the best choice at this time,
computing MoverScores incurs a cubic time complexity \cite{zhao2019moverscore}. Fortunately, this task is highly
parallelizable and we have implemented a parallel program to carry out this task on a GPU, which provides much more efficient computation of CSD-1. Nevertheless, finding a more effective and efficient measure for content similarity is highly desirable for our tasks, particularly for long articles.

%To reduce computation complexity when computing CSD-1 for a given article, our approach is to choose a moderate fixed number of text blocks at random with a moderate block size. While this approach has produced good results, it would be interesting to explore if this can be done in a different way to obtain better results.

We would also like to seek intuitions and mathematical explanations why the functions $LC_x(a,b \mid \alpha, \beta)$ resemble CSD-1 curves. 

Finally, we would like to explore if CSDs may be used to assess the overall quality of an article with a single score with better accuracy than
an early attempt \cite{YongjieWang2022} using a multi-scale essay representation that can be jointly learned, which employs multiple losses and transfer learning from out-of-domain essays.

\subsection*{Acknowledgment}
We would like to thank Jay Belanger for a valuable suggestion on function transformation.

\bibliographystyle{apalike}
{\small
\bibliography{KDIR2023v4/CSD}}

\end{document}